 \documentclass[preprint,review,3p,12pt]{elsarticle}

\usepackage{caption}
\usepackage{subcaption}
\usepackage{bbding}
\usepackage{utfsym}
\usepackage{multirow}
\usepackage{graphicx}
\usepackage[graphicx]{realboxes}
\usepackage{amsmath}
\usepackage{amsfonts}

\usepackage{enumitem}
\usepackage{newfloat}
\usepackage{listings}

\journal{Elsevier }

\begin{document}

\begin{frontmatter}

\title{Unified Image Restoration and Enhancement: Degradation Calibrated Cycle Reconstruction Diffusion Model}

\author[1]{Minglong Xue} 
\author[1]{Jinhong He}
\author[2]{Shivakumara Palaiahnakote}
\author[3]{Mingliang Zhou}
\affiliation[1]{organization={College of Computer Science and Engineering},
            addressline={Chongqing University of Technology}, 
            city={Chongqing University of Technology},
            postcode={400054}, 
            country={China}}

\affiliation[2]{organization={School of Science, Engineering and Environment},
            addressline={ University of Salford}, 
            city={Manchester},
            country={UK}}
\affiliation[3]{organization={College of Computer Science},
            addressline={Chongqing University}, 
            city={Chongqing},
            postcode={400044}, 
            country={China}}
\begin{abstract}
Image restoration and enhancement are pivotal for numerous computer vision applications, yet unifying these tasks efficiently remains a significant challenge. Inspired by the iterative refinement capabilities of diffusion models, we propose CycleRDM, a novel framework designed to unify restoration and enhancement tasks while achieving high-quality mapping. Specifically, CycleRDM first learns the mapping relationships among the degraded domain, the rough normal domain, and the normal domain through a two-stage diffusion inference process. Subsequently, we transfer the final calibration process to the wavelet low-frequency domain using discrete wavelet transform, performing fine-grained calibration from a frequency domain perspective by leveraging task-specific frequency spaces. To improve restoration quality, we design a feature gain module for the decomposed wavelet high-frequency domain to eliminate redundant features. Additionally, we employ multimodal textual prompts and Fourier transform to drive stable denoising and reduce randomness during the inference process. After extensive validation, CycleRDM can be effectively generalized to a wide range of image restoration and enhancement tasks while requiring only a small number of training samples to be significantly superior on various benchmarks of reconstruction quality and perceptual quality. The source code will be available at https://github.com/hejh8/CycleRDM.
\end{abstract}
\begin{keyword}
Image enhancement, Image restoration, Diffusion model.


\end{keyword}

\end{frontmatter}


\begin{figure*}[ht!]
\centering
  \includegraphics[height=0.5\linewidth,width=\linewidth]{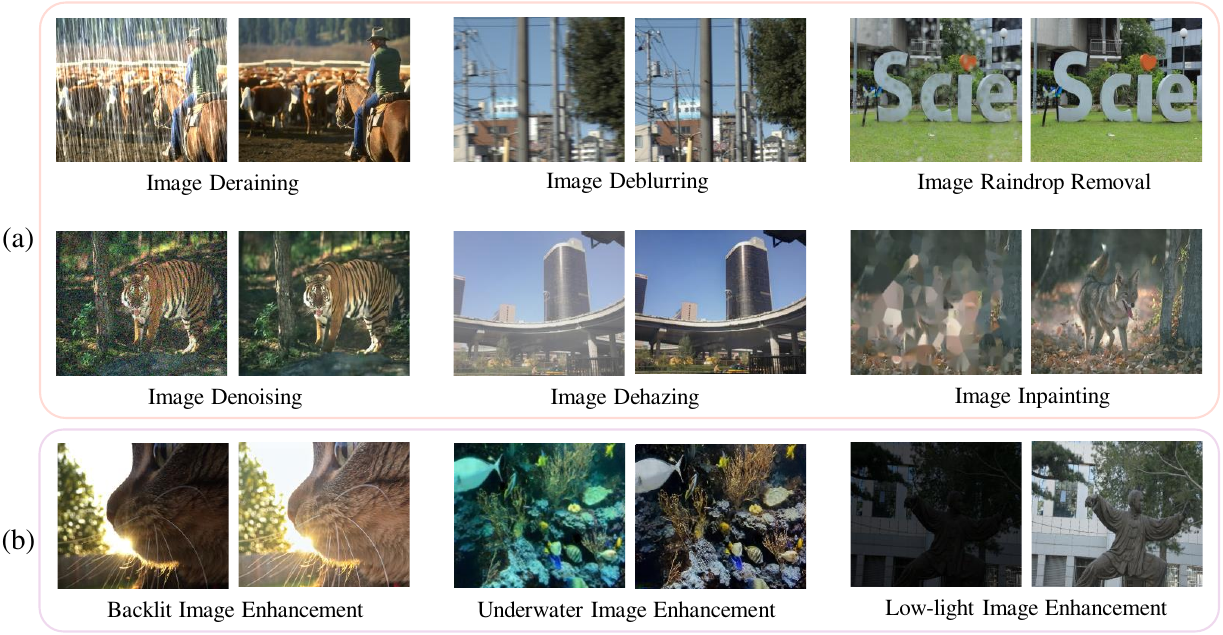}
  \caption{CycleRDM is capable of generating high-fidelity restoration in a variety of tasks. CycleRDM gives faithful results on a wide range of {\bf (a)} linear image restoration tasks. Meanwhile, CycleRDM also realizes {\bf (b)} blind, non-linear image enhancement tasks with high quality.}
  \label{fig:1}
\end{figure*}
\section{Introduction}
\label{sec1}

Complex environmental conditions in the real world usually cause damage to the quality of captured images, leading to performance degradation in various computer vision applications, such as object detection \cite{yu2024degradation}. Image restoration and enhancement aims to reverse the various degraded domains and restore the original clean image by learning the mapping relationship between the degraded and normal domains, thus improving the performance of various downstream tasks. In practice, most image restoration tasks are usually solved as linear inverse problems, such as image dehazing \cite{qin2020ffa,tu2022maxim}, image deraining \cite{xiao2022image,zhang2024prior}, and image deblurring \cite{park2020multi,mao2023intriguing}, where the degradation model is usually linear and known. However, image enhancement tasks are more often studied as nonlinear blind problems, such as image low-light enhancement \cite{xu2022snr,xu2025upt}, underwater image enhancement \cite{cong2023pugan}, and backlight image enhancement \cite{liang2023iterative}, where we need to simultaneously estimate the degradation model and recover clean images with high fidelity. Therefore, it is challenging to try to effectively unify the two types of tasks.


Along with the rise of deep learning, numerous researchers have been bridging the gap between degraded and normal images by learning powerful prior knowledge in large-scale datasets through a data-driven approach \cite{li2022all, luo2024controlling,xu2022snr,li2023embedding}. However, existing advances are still challenged by the lack of ability to model various complex degradation conditions in the real world. Therefore, there has been a surge of interest in seeking broader prior knowledge of images through generative models \cite{yang2023implicit,yang2023robust,cong2023pugan}. Especially, recent diffusion models have received much attention (such as image restoration \cite{fei2023generative,lugmayr2022repaint,luo2023image} and enhancement \cite{jiang2023low,hou2024global,lv2024fourier,jiang2024lightendiffusion}) for their impressive performance in image generation tasks. Specifically, diffusion models are trained to iteratively denoise the image by reversing a diffusion process to achieve a mapping from randomly sampled Gaussian noise to a complex target distribution without suffering from model collapse as GANs.

As a class of likelihood-based models, diffusion models model the details of the data through a large number of inference time steps in image generation tasks. For image restoration and enhancement tasks, we only need to repair the degraded regions in the given corrupted image, and thus adopting the image generation paradigm may lead to the generation of mismatched image details. Therefore, most of the existing methods \cite{hou2024global,jiang2023low,huang2024wavedm} use fewer time steps (about 10 - 50 steps) to bridge the gap in the degraded domain through single-stage inference. However, the various degradation factors arising from complex real-world environments make it challenging to construct high-quality mappings with single-stage inference. This also complicates finding an efficient balance between linear degradation and blindness problems, leading to difficulties in developing high-quality, unified models for image restoration and enhancement tasks.

In this paper, we propose a Cycle Reconstruction Diffusion Model (CycleRDM) that performs fine-grained calibration of the degraded domain by employing a multi-stage diffusion inference process to achieve high-quality mappings from degraded images to normal images. This approach ensures the stable unification of image restoration and enhancement tasks. Specifically, we first learn the mapping relationships between the degraded domain to the rough normal domain and the rough normal domain to the fine normal domain through a two-stage diffusion inference process. Subsequently, we use the prior learned knowledge to guide the final calibration process. In particular, inspired by \cite{huang2024wavedm,jiang2023low}, we transfer the final calibration process to the wavelet low-frequency domain to achieve high-quality domain mapping from the frequency domain perspective and alleviate the consumption of computational resources. Meanwhile, we design a feature gain module using residual dense blocks to further remove redundant features from the wavelet high-frequency information in restoration tasks. To further stabilize the denoising process and minimize randomness during inference, we explore the use of multimodal text and Fourier transform-driven appearance reconstruction. As shown in Fig. 1, CycleRDM effectively unifies nine different types of degradation and produces visually appealing results. Extensive experiments show that CycleRDM achieves highly competitive performance, even with a limited amount of training data. Our main contributions are summarized below:
\begin{itemize}[noitemsep,topsep=10pt]
\item We propose CycleRDM, which exploits the generative power of the diffusion model and combines it with the wavelet transform to construct a novel multi-stage diffusion inference process that achieves fine calibration and high-quality mapping of the degenerate domain using only a small amount of training data.
\item We designed a feature gain module using residual dense blocks to further remove redundant features from the wavelet high-frequency information. In addition, we drive the reconstruction of the appearance and improve the stability of the inference process by combining multimodal text and Fourier transform.
\item Through extensive experiments conducted on tasks involving nine different types of degradation, we demonstrate the effectiveness of CycleRDM. Our approach efficiently unifies image restoration and enhancement tasks, delivering highly competitive performance.
\end{itemize}

\section{Related work}

\subsection{Image Restoration}
Image restoration aims to achieve high-quality mapping of degraded images to normal images, which is a long-standing problem in computer vision and encompasses a variety of tasks such as image denoising \cite{wu2020unpaired,tian2023multi}, deraining \cite{zhang2024prior,xiao2022image}, dehazing \cite{qin2020ffa,yang2022self,sun2024unsupervised}, deblurring \cite{li2022learning,mao2023intriguing}. Most of the existing methods focus on some single degradation task, using huge data-driven construction of high-quality mapping relations. However, recently, unified image restoration research has also begun to receive widespread attention \cite{li2022all,chen2022simple}. For example, NAFNet \cite{chen2022simple} derives a nonlinear activation-free network for multiple image restoration tasks by exploring the necessity of nonlinear activation functions. AIRNet \cite{li2022all} exploits the consistency of the same degraded image and the inconsistency present in different degradations to learn degradation representations for robust multi-task restoration. In addition, diffusion models \cite{fei2023generative,luo2023image,huang2024wavedm} have also been widely used with impressive results due to their powerful generative capabilities. For example, IR-SDE \cite{luo2023image} achieves good recovery results by constructing stochastic differential equations to reverse image degradation. WaveDM \cite{huang2024wavedm} Learning clean image distributions in the wavelet domain by performing wavelet transform embedded diffusion model to reduce the inference time overhead and achieved effective image recovery. However, despite their successes, these models still face limitations due to the lack of prior knowledge and the difficulty in modeling unknown degradation, which restricts them to linear image restoration tasks and limits their applicability to more complex, blind restoration problems.


\subsection{Image Enhancement}

Unlike most linear image restoration tasks, image enhancement tasks suffer from unknown degradation factors \cite{fei2023generative,luo2024controlling}. Thus solving the blind inverse problem is not trivial, as one would need to simultaneously estimate the degradation model and recover a clean image with high fidelity. To deal with these unknown degradation tasks, numerous corresponding studies have been developed, such as low-light image enhancement \cite{xu2022snr,xu2025upt,li2023embedding}, underwater image enhancement \cite{cong2023pugan,huang2023contrastive}, and backlit image enhancement \cite{liang2023iterative}. Motivated by the power of the latest diffusion models in capturing and modifying data distributions and features, researchers have gradually explored the potential of generative paradigms to bridge the gap between unknown degradation domains \cite{lv2024fourier,jiang2023low,cong2023pugan}. For example, GSAD \cite{hou2024global} leads to robust and effective low-light enhancement by establishing a global regularisation embedded in the diffusion process. LightenDiffusion \cite{jiang2024lightendiffusion} performs self-consistent enhancement effects by incorporating Retinex theory in the diffusion latent space. GDP \cite{fei2023generative} proposed to use diffusion priors to generate realistic output and use this to connect image restoration and enhancement tasks. However, due to model limitations, these methods suffer from significant time overhead while failing to strike a stable balance between linear and blind problems, resulting in still task-biased model performance. In addition, recent DA-CLIP \cite{luo2024controlling} utilizes image controllers to predict degradation and adjust a fixed CLIP image encoder to unify image restoration and enhancement, but it still rely on large amounts of training data and face challenges in generating high-quality images.

\begin{figure*}[t]
        \centering
        \includegraphics[height=0.55\linewidth,width=\linewidth]{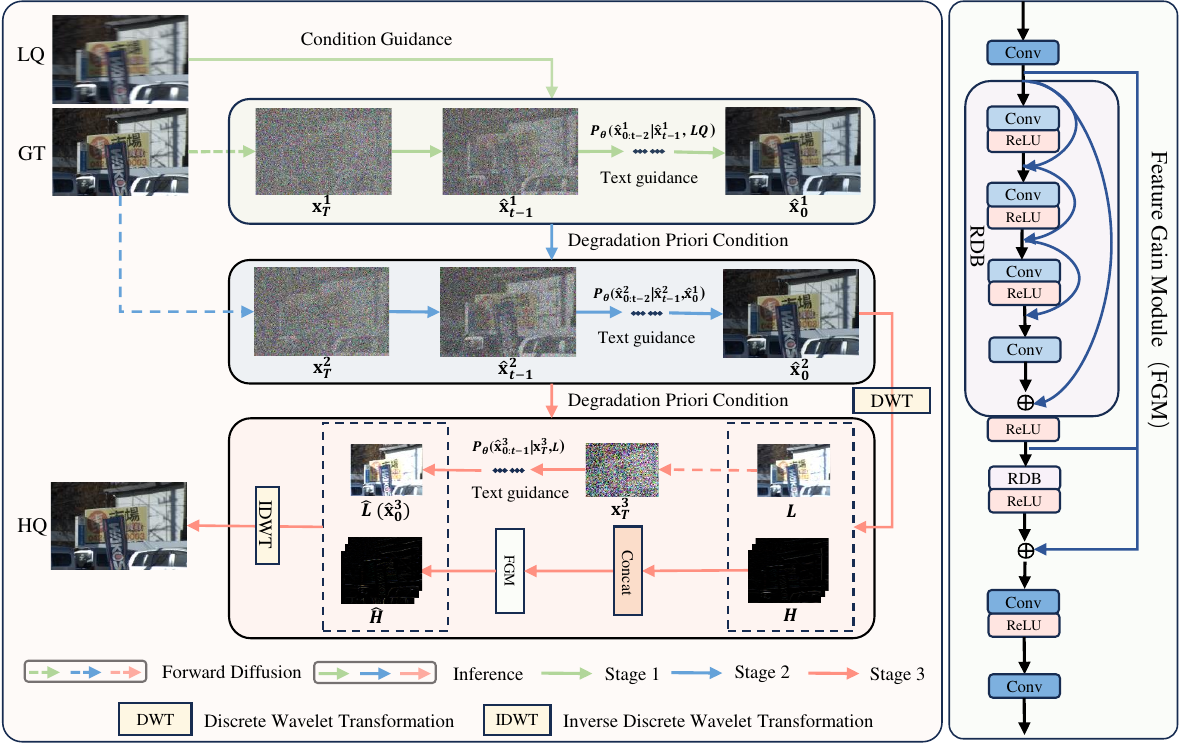}
        \caption{The proposed overall framework for CycleRDM. We use image deblurring as a demonstration. Firstly, in Stage 1 we use the degraded image $LQ$ as a condition guidance to learn the mapping relation between the degraded domain to the rough normal domain, and later to guide the learning of the rough normal domain to the normal domain in Stage 2. In Stage 3, we perform a discrete wavelet transform ($DWT$) on the output $\hat{x}^{2}_{0}$ of stage 2. At the same time, a fine calibration is performed in the wavelet low-frequency domain $L$ using the degradation prior learned earlier. For each Stage output, we also utilise multimodal text for appearance guidance. And the high-frequency $H$ is enhanced by the feature gain module ($FGM$), and finally recovered to a high-quality image $HQ$ by the inverse discrete wavelet transform ($IDWT$).}
        \label{fig:2}
    \end{figure*}
\section{METHODOLOGY}

In this study, we aim to explore diffusion models with fine calibration of degraded domains to unify image restoration and enhancement tasks effectively, significantly reduce the performance bias when the task is expanded, and enhance the stability of the model effect. Specifically, as shown in Fig. \ref{fig:2}, we use deblurring as a demonstration example, and CycleRDM uses a three-stage diffusion inference process to gradually learn the mapping from a degraded domain to a normal domain with fine calibration. In addition, we further refine the features by designing a feature gain module to remove redundant features from the wavelet high-frequency information. Finally, multimodal text and the Fourier frequency domain are used to drive appearance reconstruction further and reduce content randomness in the inference process. In the following sections, we will elaborate on the proposed method.

\subsection{Diffusion models Preliminary} \label{Preliminary}

Diffusion models to train Markov chains by variational inference. It converts complex data into completely random data by adding noise and gradually predicts the noise to recover the expected clean image. Consequently, it usually includes the forward diffusion process and reverse inference process.

The forward diffusion process mainly relies on incremental introduction of Gaussian noise with fixed variance $\{\beta_t\in(0, I)\}_{t=1}^T$ into the input distribution ${x}_0$ until the time steps of T approximate purely noisy data. This process can be expressed as:

\begin{equation}
q(x_t|x_{t-1})=N(x_t;\sqrt{1-\beta_t}x_{t-1},\beta_tI),
\tag{1}
\end{equation}
where $x_t$ and $\beta_t$ are the corrupted noise data and the predefined variance at time step $t$. Respectively, $N$ denotes a Gaussian distribution. Furthermore, each time step $x_t$ of the forward diffusion process can be obtained directly by computing $x_0$:
\begin{equation}\label{2}
x_t=\sqrt{\overline{\alpha}_t}x_0+\sqrt{1-\overline{\alpha}_t}\epsilon,\ \  \epsilon \sim N(0, I),
\tag{2}
\end{equation}
where $\alpha_t=1-\beta_t$, $\overline{\alpha}_t$=$\prod_{i=1}^{t}\alpha_i$.

The reverse inference process is to recover the original data from Gaussian noise. In contrast to the forward diffusion process, the reverse inference process relies on optimizing the noise predictor to iteratively remove the noise and recover the data until the randomly sampled noise $x_{T}\sim N(0, I)$ becomes clean data $\hat{x}_0$. Formulated as:
\begin{equation}\label{3}
p_\theta(\hat{x}_{t-1}|\hat{x}_t)=N(\hat{x}_{t-1};\mu _\theta(\hat{x}_t,t),\sigma ^2_tI),
\tag{3}
\end{equation}
where $\mu _\theta$ is the diffusion model noise predictor, which is mainly optimized by the editing and data synthesis functions and used as a way to learn the denoising process, as follows:
\begin{equation}
\mu _\theta=\frac{1}{\sqrt{\alpha_t}}(x_t-\frac{\beta_t}{\sqrt{1-\overline{\alpha}_t}}\epsilon_\theta(x_t,t)),
\tag{4}
\end{equation}
where $\epsilon_\theta$ is a function approximator intended to predict noise vectors $\epsilon$ from $\hat{x}_t$.

\subsection{Multi-Stage Diffusion Inference Process}
In the single-stage denoising process of the diffusion model, the content diversity caused by randomly sampled noise is undesirable for image restoration and enhancement tasks, which will lead to instability in their performance. To address this, we propose stabilizing the model by gradually learning the gap between the degraded and normal domains through a multi-stage process. Additionally, we guide the calibration process using the learned degraded prior, which helps ensure more consistent and reliable performance.


Concretely, we first input the low-quality degraded image $LQ$ as a condition in Stage $1$, and map the degraded domain into the rough normal domain through diffusion inference. Where a Prior knowledge of some of the degraded parameters is stored through the initially recovered image $\hat{x}^{1}_{0}$. Subsequently, in Stage $2$, we use the $\hat{x}^{1}_{0}$ as a conditional input to further bridge the gap between the rough normal domain and the normal domain using the degradation parameters learned in Stage $1$. Where, as inputs are made at each stage, we simultaneously use multimodal text to guide them through the stages. Especially in stage $1$ and stage $2$, we only set the time step of the forward diffusion process to $200$. Meanwhile, the inverse denoising distribution is rewritten as conditional distribution: 
\begin{equation}
p_\theta(\hat{x}_{0:T}|y)=p(\hat{x}_T)\prod_{t=1}^{T} p_\theta(\hat{x}_{t-1}|\hat{x}_t,y),
\tag{5}
\end{equation}
where $y$ is the input image condition. The function approximator $\epsilon_\theta(x_t,t)$ becomes $\epsilon_\theta(x_t,t,y)$.

Finally, we use the output $\hat{x}^{2}_{0}$ in Stage 2 as a corrupted observation of the normal image $GT$. Furthermore, inspired by \cite{jiang2023low}, we transfer Stage 3 to the low-frequency wavelet domain to calibrate the degradation and to alleviate the consumption of computational resources. Specifically, we employ the discrete wavelet transform $DWT(\cdot)$ which decomposes $\hat{x}^{2}_{0}$ into low-frequency information $L$ and high-frequency information $H$. 
\begin{equation}
\{L,H\}={ DWT}(\hat{x}^{2}_{0}),
\tag{6}
\end{equation}
where $L$ contains the main information about the structure of the image content and $H$ contains three high-frequency subbands in vertical, horizontal and diagonal directions.

In addition, due to the characteristics (Eq. \ref{2}) of the forward diffusion process, we perform diffusion and inference simultaneously in all three stages. Notably, we can perform fewer forward diffusion time steps in the third stage to obtain more conditional priors. With the prior degradation learned in the previous stages, $L$ a refined calibration fine-tuning of 10 steps in the wavelet domain removes redundant information, resulting in a high-quality mapping. Thus, the overall optimization objective of the denoising network can be formulated as:
\begin{equation}
\mathcal{L}_{diff}=\sum_{s=1}^{3}\tau_s \sum_{t=0}^{T} \mathbb{E}_s\left [ \parallel \epsilon_t-\epsilon_\theta(x_t,t,y)\parallel_2\right ],
\tag{7}
\end{equation}
where $s$ is the stage of the diffusion inference process. $\tau_s$ is the weight parameter optimized by the noise reducer at each stage, which we set to $\tau_{s=1, 2}=1$ and $\tau_{s=3}=0.9$.

\subsection{Feature Gain Module}
To further obtain the image restoration task performance, a feature gain module is designed for the wavelet high-frequency information decomposed by the calibration process, as shown in Fig. \ref{fig:2}. Redundant features are further removed by combining the residual dense blocks (RDB) \cite{tian2023multi} and wavelet transform. Specifically, we first extract shallow features through a convolutional layer with a convolutional kernel size of $5 \times 5$. Subsequently, a combination of 4-layers of RDB and ReLU was used to refine the features in the image further. At the same time, we fused the extracted features by two residual operations to enhance the memory ability of shallow features to deeper ones to prevent long-term dependency. Finally, we again refine the extracted features and reconstruct the noise mapping by a combination of convolutional layers and ReLU. By removing the learned noise mappings, we further reconstruct $H$ into clean high-frequency information $\hat{H}$. In this paper, the number of input channels and output channels of both RDBs is 64, the number of input channels of the last convolutional layer is 64, and the number of output channels is 3. Meanwhile, based on the enhanced high-frequency information $\hat{H}$ and calibrated low-frequency information $\hat{L}$, we transform it into the final high quality recovered image $HQ$ by inverse discrete wavelet transform $IDWT(\cdot)$:
\begin{equation}
HQ=IDWT(\hat{H},\hat{L}).
\tag{8}
\end{equation}

\subsection{Appearance Guidance And Network Training}
In CycleRDM, in addition to the objective function $\mathcal{L}_{diff}$ used to optimize the diffusion model, We also utilize multimodal text combined with a multi-stage process to construct a multi-level semantic guidance network that drives appearance reconstruction and process inputs at each stage through a frozen CLIP model. Specifically, for the outputs of each inference stage $\sum_{s=1}^{3}\hat{x}_0^s$, we encode textual prompts by input pairing, which promotes recovery results close to positive prompts and away from negative prompts by performing computing semantic similarities in the CLIP latent space. Compared to most existing methods that use image-level supervised outputs, we supervise image generation at both semantic and appearance levels separately, effectively bridging the gap between metric-favorable and visually friendly and reducing model training obfuscation. We can represent this as:
 
\begin{equation}
\mathcal{L}_{clip}=\sum_{s=1}^{3}\frac{e^{\cos(\Phi_{image} (\hat{x}_0^s ),\Phi_{text}(T_n ))} }{\sum _{j\in\{T_p,T_n\}} e^{\cos(\Phi_{image} (\hat{x}_0^s),\Phi_{text}(T_j))}},
\tag{9}
\end{equation}
where $T_p$ denotes positive prompts text (such as high light images), $T_n$ denotes negative prompts text (such as low light images), $\Phi_{text}$ denotes text encoder, and $\Phi_{image}$ denotes image encoder. Furthermore, we utilize a content loss $\mathcal{L}_{content}$ that combines the MSE loss and the SSIM loss, in order to minimize the content difference between the recovered image and the normal image:
\begin{equation}
     \begin{split}
\mathcal{L}_{content}=&\sum_{s=1}^{3}\sum_{l=0}^{4}{\omega_l\parallel\Phi_{image}^l(\hat{x}_0^s)- \Phi_{image}^l(GT)\parallel_2}\\
&+(1-SSIM(HQ,GT)),
     \end{split}
\tag{10}
\end{equation}
where $\omega_l$ is the weight of layer l of the image encoder in the ResNet101 CLIP model. We set $\omega_{l=0,1,2,3}=1$ and $\omega_{l=4}=0.5$. Meanwhile, we introduce the frequency-aware loss $\mathcal{L}_{fre}$ to learn the GT spectrum, which consists of two components: amplitude $amp$ and phase $pha$:
\begin{equation}
amp_{hq},pha_{hq}=\mathcal{F}_{df}(HQ),
\tag{11}
\end{equation}
\begin{equation}
amp_{gt},pha_{gt}=\mathcal{F}_{df}(GT),
\tag{12}
\end{equation}
\begin{equation}
\mathcal{L}_{fre}=\vartheta_1\parallel{amp_{hq}-amp_{gt}\parallel_1}+\vartheta_2\parallel{pha_{hq}-pha_{gt}\parallel_1},
\tag{13}
\end{equation}
where $\mathcal{F}_{df}$ denotes the fast Fourier transform (FFT), $\vartheta_1$ and $\vartheta_2$ are the weighting parameters for amplitude loss and phase loss and were all set to 0.5 based on experience.

Thus, the total training loss can be defined as:
\begin{equation}
\mathcal{L}_{total}=\mathcal{L}_{diff}+\gamma_1\mathcal{L}_{clip}+\mathcal{L}_{content}+\gamma_2\mathcal{L}_{fre},
\tag{14}
\end{equation}
where $\gamma_1$ and $\gamma_2$ are hyperparameters set to 0.2 and 0.3, respectively.

\section{EXPERIMENTS}
In this section, we describe the implementation details, datasets in detail. We validate the generalization of CycleRDM by performing a systematic comparison of CycleRDM on a series of different tasks after training. In particular, compared to other baselines, we select only a small number of training samples of no more than 500 in each different dataset. We also conduct a series of ablation experiments to validate the effectiveness of the proposed design. 
\subsection{Datasets}
We provide detailed information about the dataset for all tasks. For each task's dataset, as shown in Table \ref{111}, we select only a small number of random images (up to 500) for training and test them uniformly on the test set. Compared to the baseline DA-CLIP and IR-SDE we greatly alleviate the training data and achieve superior performance. Therefore, for each task, in addition to the two baselines of unified image recovery methods, we compare more with the state-of-the-art methods for the particular task. The details are as follows:
\begin{table*}[t]
\caption{Details of the number of images we selected for training and testing for each task.}
\scalebox{0.66}{
\begin{tabular}{cccccccccc}
\hline
Image Number & Dehazing & Deraining & Deblurring & Denoising & Raindrop Remova & Inpainting & Low-light & Underwater & Backlight \\ \hline
Train Phase  & 500                              & 500                               & 500                                & 500                               & 500                                     & 500                                & 485                               & 500                                & 0                                 \\ \hline
Test Phase   & 1000                             & 100                               & 50                                 & 68                                & 58                                      & 100                                & 170                               & 50                                 & 30                                \\ \hline
\end{tabular}
}
\label{111}
\end{table*}
\begin{itemize}[noitemsep,topsep=10pt]
\item {\bf Image Deraining:} We use the Rain100H dataset \cite{yang2017deep} for training and testing. This is a synthetic dataset containing 1800 paired training images and 100 test images. Among them, We randomly selected 500 images to add to the training data and compared them on 100 test sets.
\item {\bf Image Dehazing:} We use the RESIDE-6k dataset \cite{qin2020ffa} for training and testing, which is a mixture of indoor and outdoor images, with 6000 images for training and 1000 images for testing. In particular, we select only a small number of 500 images for training.
\item {\bf Image Denoising:} We trained on 500 randomly selected images from other task datasets, but all LQ images were generated by adding Gaussian noise at noise level 50. The test images are from CBSD68 \cite{martin2001database}, a dataset with 68 denoised images and Gaussian noise added.
\item {\bf Image Deblurring:} We use the GoPro for training as other methods and validate on the BSD dataset \cite{zhong2023real}. In this case, the training images are still only 500, and we randomly select 50 from the BSD dataset for the test images.
\item {\bf Image Raindrop Removal:} We use the RainDrop dataset \cite{qian2018attentive}, which contains 861 images for training and 58 images for testing.
\item {\bf Image Inpainting:} We use 256-resolution images from CelebaHQ as the training dataset, which contains 30,000 images, and We used 500 images to add to the training data and 100 images segmented using 100 thin masks in RePaint \cite{lugmayr2022repaint} for testing. In addition, we also take a random masking process for some of the images used for testing.
\item {\bf Low-light Image Enhancement:} We use the LOLv1 dataset \cite{wei2018deep} for training, which contains 485 paired images for training and 15 images for testing. Also, we test directly on the LOLv2-real dataset \cite{yang2021sparse}, which contains 100 test images. In addition, we test on the unpaired datasets LIME \cite{guo2016lime} and DICM \cite{lee2013contrast}.
\item {\bf Backlight Image Enhancement:}  We randomly selected 30 images from the BackLit300 dataset \cite{liang2023iterative} for testing.
\item {\bf UnderWater Image Enhancement:} We use the LSUI dataset \cite{peng2023u}, consisting of 5004 image pairs, which involves richer underwater scenes (lighting conditions, water types, and target classes) and better visual quality reference images than the existing ones. The test data were randomly selected from 50 images, and the training images were randomly selected from 500 images.
\end{itemize}

\subsection{Implementation Details}
We implemented our method on two NVIDIA Tesla V100s GPUs using PyTorch on an Ubuntu system. We performed a total of 1000 epochs on the network using the Adam optimizer, with the initial learning rate set to $1\times10^{-4}$, and the patch size set to $128\times128$ respectively. To achieve efficient recovery, In order to achieve efficient recovery,  time steps were set to 200 for the training phase, and for the denoising phase, the implicit sampling step was set to 10.

\subsection{Image Restoration Tasks}
To quantify the performance of CycleRDM, we show the evaluated effects of four specific degradation tasks: image deraining on the Rain100H dataset \cite{yang2017deep}; image denoising on the CBSD68 dataset \cite{martin2001database}; image deblurring on the BSD dataset \cite{zhong2023real}; image dehazing on the RESIDE-6K dataset \cite{qin2020ffa}; image raindrop removal on the RainDrop dataset \cite{qian2018attentive}; and image inpainting on the CelebaHQ dataset. For each experiment, we measure the fidelity of recovery performance and the quality of the resulting images using two distortion metrics, PSNR and SSIM \cite{wang2004image}, and two perceptual metrics, LPIPS \cite{zhang2018unreasonable} and FID \cite{heusel2017gans}. 

\begin{table*}[t!]
\renewcommand\arraystretch{1.2}
    \centering
        \caption{Quantitative comparison of our method with other state-of-the-art methods between four different image restoration tasks. The best and second performance are marked in {\textcolor[HTML]{FF0000}{red}} and {\color[HTML]{00B0F0}{blue}}, respectively. }
    \begin{subtable}[t]{0.45\linewidth}
    \caption{Dehazing Quantitative comparison  on the RESIDE-6k dataset.}
\resizebox{!}{2.96cm}{
\begin{tabular}{l|c|cccc}
\hline
                          &                             & \multicolumn{4}{c}{RESIDE-6k}                                                                                               \\ \cline{3-6} 
\multirow{-2}{*}{Methods} & \multirow{-2}{*}{Reference} & PSNR↑                          & SSIM↑                         & LPIPS↓                        & FID↓                          \\ \hline
FFANet \cite{qin2020ffa}                    & AAAI'20                     & 20.766                        & 0.891                        & 0.084                        & 9.112                        \\
MSFNet \cite{zhu2021multi}                   & TIP'21                      & 23.931                        & 0.914                        & 0.071                        & 10.166                       \\
MAXIM  \cite{tu2022maxim}                   & CVPR'22                     & 29.121                        & 0.932                        & 0.045                        & 8.116                        \\
IR-SDE \cite{luo2023image}                   & ICML'23                     & 25.250                        & 0.908                        & 0.062                        & 8.330                        \\
D4\_plus \cite{yang2023robust}                      & IJCV'23                     & 25.875                        & 0.926                        & 0.037                        & 9.826                        \\
UME-Net  \cite{sun2024unsupervised}                  & PR'24                     & 26.766                        & {\color[HTML]{00B0F0} 0.938 }                       & 0.041                        & 8.395                        \\
DA-CLIP \cite{luo2024controlling}                  & ICLR'24                     & {\color[HTML]{FF0000} 30.062} & 0.935                        & {\color[HTML]{00B0F0} 0.033} & {\color[HTML]{FF0000} 5.341} \\
Ours                      & -                         & {\color[HTML]{00B0F0} 29.202} & {\color[HTML]{FF0000} 0.960} & {\color[HTML]{FF0000} 0.026} & {\color[HTML]{00B0F0} 7.095} \\ \hline
\end{tabular}
}
    \end{subtable}
    \begin{subtable}[t]{0.45\linewidth}
    \caption{Denoising Quantitative comparison on the CBSD68 dataset.}
\resizebox{!}{2.96cm}{
\begin{tabular}{l|c|cccc}
\hline
                          &                             & \multicolumn{4}{c}{CBSD68}                                                                                                  \\ \cline{3-6} 
\multirow{-2}{*}{Methods} & \multirow{-2}{*}{Reference} & PSNR↑                          & SSIM↑                         & LPIPS↓                        & FID↓                           \\ \hline
DBSN \cite{wu2020unpaired}                     & ECCV'20                     & 26.456                        & 0.713                        & 0.281                        & 107.972                       \\
DCDicL \cite{zheng2021deep}                    & CVPR'21                     & {\color[HTML]{FF0000} 28.565} & {\color[HTML]{00B0F0} 0.786} & 0.235                        & 83.128                        \\
Restormer \cite{zamir2022restormer}                & CVPR'22                     & 27.246                        & 0.762                        & {\color[HTML]{00B0F0} 0.215} & 87.140                        \\
AirNet \cite{li2022all}                     & CVPR'22                     & {\color[HTML]{00B0F0}27.511} & 0.769                        & 0.264                        & 93.890                        \\
IR-SDE \cite{luo2023image}                     & ICML'23                     & 24.821                        & 0.640                        & 0.232                        & 79.380                        \\
WCDM \cite{jiang2023low}                     & TOG'23                      & 25.899                        & 0.699                             & 0.292                             & 103.488                              \\
DA-CLIP \cite{luo2024controlling}                   & ICLR'24                     & 24.333                        & 0.571                        & 0.269                        & {\color[HTML]{00B0F0} 69.908} \\
Ours                      & -                            & 27.424                        & {\color[HTML]{FF0000} 0.789} & {\color[HTML]{FF0000} 0.166} & {\color[HTML]{FF0000} 55.947} \\ \hline
\end{tabular}
}
    \end{subtable}
        \begin{subtable}[t]{0.45\linewidth}
        \caption{Deblurring Quantitative comparison on the BSD dataset.}
        \resizebox{!}{2.95cm}{
\begin{tabular}{l|c|cccc}
\hline
                                &                             & \multicolumn{4}{c}{BSD}                                                                                                     \\ \cline{3-6} 
\multirow{-2}{*}{Methods}       & \multirow{-2}{*}{Reference} & PSNR↑                          & SSIM↑                         & LPIPS↓                        & FID↓                           \\ \hline
MTRNN \cite{park2020multi}                          & ECCV'20                     & 25.960                        & 0.831                        & 0.195                        & 68.064                        \\
{\color[HTML]{1F2328} DID-Anet} \cite{ma2021defocus} & TIP'21                      & 25.573                        & 0.803                        & 0.243                        & 71.122                        \\
MSDI-Net \cite{li2022learning}                       & ECCV'22                     & 27.049                        & 0.859                        & 0.145                        & 49.351                        \\
NAFNet \cite{chen2022simple}                         & ECCV'22                     & 25.336                        & 0.797                        & 0.266                        & 69.268                        \\
IR-SDE \cite{luo2023image}                          & ICML'23                     & 23.723                        & 0.806                        & 0.179                        & 64.506                        \\
DeepRFT \cite{mao2023intriguing}                        & AAAI'23                     & {\color[HTML]{00B0F0} 28.995} & {\color[HTML]{FF0000} 0.901} & {\color[HTML]{00B0F0} 0.143} & {\color[HTML]{00B0F0} 47.421} \\
DA-CLIP \cite{luo2024controlling}                          & ICLR'24                     & 25.631                        & 0.812                        & 0.156                        & 53.660                        \\
Ours                        & -                             & {\color[HTML]{FF0000} 29.056} & {\color[HTML]{00B0F0} 0.889} & {\color[HTML]{FF0000} 0.113} & {\color[HTML]{FF0000} 38.643} \\ \hline
\end{tabular}
}
    \end{subtable}
        \begin{subtable}[t]{0.45\linewidth}
         \caption{Deraining Quantitative comparison  on the Rain100H dataset.}
\resizebox{!}{2.95cm}{
\begin{tabular}{l|c|llll}
\hline
                          &                             & \multicolumn{4}{c}{Rain100H}                                                                                                \\ \cline{3-6} 
\multirow{-2}{*}{Methods} & \multirow{-2}{*}{Reference} & \multicolumn{1}{c}{PSNR↑}      & \multicolumn{1}{c}{SSIM↑}     & \multicolumn{1}{c}{LPIPS↓}    & \multicolumn{1}{c}{FID↓}       \\ \hline
RecDerain \cite{ren2020single}                & TIP'20                      & 29.016                        & 0.899                        & 0.104                        & 45.698                        \\
MPRNet \cite{zamir2021multi}                   & CVPR'21                     & 28.519                        & 0.872                        & 0.152                        & 62.146                        \\
NAFNet \cite{chen2022simple}                   & ECCV'22                     & 27.322                        & 0.853                        & 0.149                        & 46.350                         \\
IDT \cite{xiao2022image}                      & TPAMI'23                    & {\color[HTML]{FF0000} 32.148} & {\color[HTML]{FF0000} 0.931} & {\color[HTML]{00B0F0} 0.065} &  25.951 \\
IR-SDE  \cite{luo2023image}                       & ICML'23                     & {\color[HTML]{343434} 29.097} & 0.886                        & {\color[HTML]{FF0000} 0.050}  & {\color[HTML]{00B0F0} 20.869} \\
WSDformer \cite{zhang2024prior}                & TMM'24                      & 25.663                        & 0.775                        & 0.092                        & 26.409                        \\
DA-CLIP \cite{luo2024controlling}                  & ICLR'24                     & 28.756                        & 0.849                        & 0.071                        & 29.214                        \\
Ours                      & -                            & {\color[HTML]{00B0F0} 30.407} & {\color[HTML]{00B0F0} 0.908} & {\color[HTML]{FF0000} 0.050} & {\color[HTML]{FF0000} 20.318} \\ \hline
\end{tabular}
}
    \end{subtable}
    \label{tab:1}
\end{table*}

\begin{table}[t!]
\caption{Quantitative comparison of our method with other state-of-the-art methods in image inpainting and image raindrop removal tasks.}
\centering
\renewcommand\arraystretch{1.2}
\scalebox{0.82}{
\begin{tabular}{c|l|c|cccc}
\hline
Tasks                        & Methods   & Reference & PSNR↑                          & SSIM↑                         & LPIPS↓                        & FID↓                           \\ \hline
                             & Restormer \cite{zamir2022restormer} & CVPR'22   & 29.881 & {\color[HTML]{00B0F0} 0.914}                        & 0.068                        & 31.960                        \\
                             & IR-SDE \cite{luo2023image}    & ICML'23   & 27.557                        & 0.884                        & 0.061                        & 29.605                        \\
                             & GDP \cite{fei2023generative}      & CVPR'23   & {\color[HTML]{FF0000} 32.028} & 0.889                        & 0.046                        & 28.318                        \\
                             & DA-CLIP\cite{luo2024controlling}   & ICLR'24   & 29.277                        & 0.901 & {\color[HTML]{00B0F0} 0.042} & {\color[HTML]{00B0F0} 22.684} \\
\multirow{-5}{*}{Inpainting} & Ours      & -         & {\color[HTML]{00B0F0} 30.292} & {\color[HTML]{FF0000} 0.926} & {\color[HTML]{FF0000} 0.039} & {\color[HTML]{FF0000} 22.053} \\ \hline
                             & AirNet \cite{li2022all} & CVPR'22   & 29.085                        & {\color[HTML]{FF0000}  0.916} & 0.099                        & 38.310                        \\
                             & IR-SDE \cite{luo2023image}   & ICML'23   & 28.191                        & 0.836                        & 0.107                        & 34.221                        \\
                             & WaveDM \cite{huang2024wavedm}   & TMM'24   & 28.905                        & 0.866                        & 0.113                       & 36.321                          \\
                             & DA-CLIP \cite{luo2024controlling}  & ICLR'24   & {\color[HTML]{00B0F0} 29.305} & 0.882                        & {\color[HTML]{00B0F0} 0.061} & {\color[HTML]{FF0000} 22.379} \\
\multirow{-5}{*}{Raindrop}   & Ours      & -         & {\color[HTML]{FF0000} 29.857} & {\color[HTML]{00B0F0} 0.908} & {\color[HTML]{FF0000} 0.058} & {\color[HTML]{00B0F0} 28.879} \\ \hline
\end{tabular}
}
\label{tab:r1}
\end{table}

\begin{figure}[ht!]
        \centering
        \includegraphics[height=0.7\linewidth,width=\linewidth]{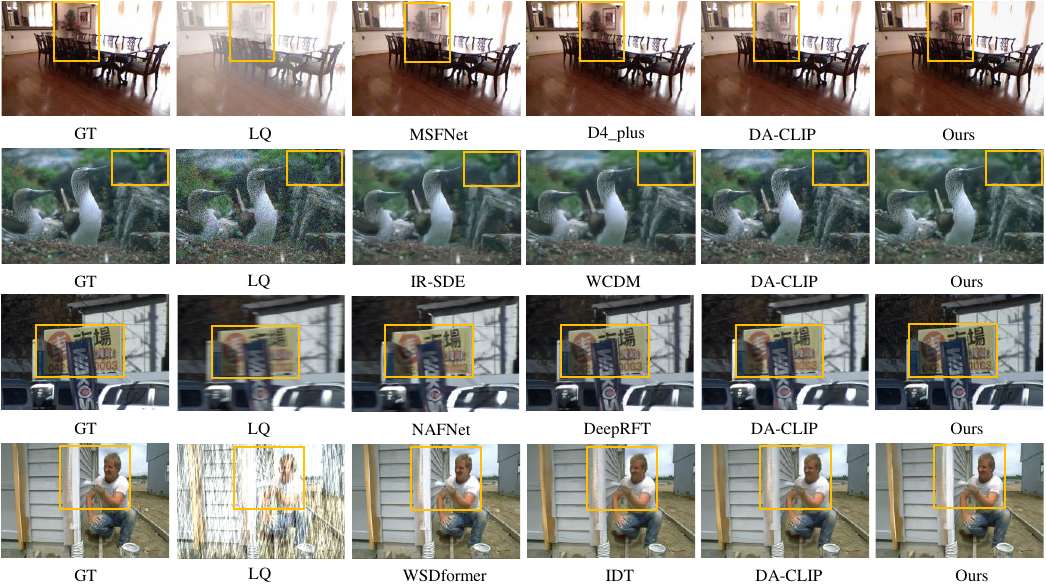}  
        \caption{Comparison of our method with other methods on 4 different degradation-specific tasks. Where the first to fourth rows are dehazing, denoising, deblurring, and deraining respectively. Best viewed by zooming in.}
        \label{fig:3}
    \end{figure}
\begin{figure*}[ht!]
        \centering
        \includegraphics[height=0.85\textwidth,width=\textwidth]{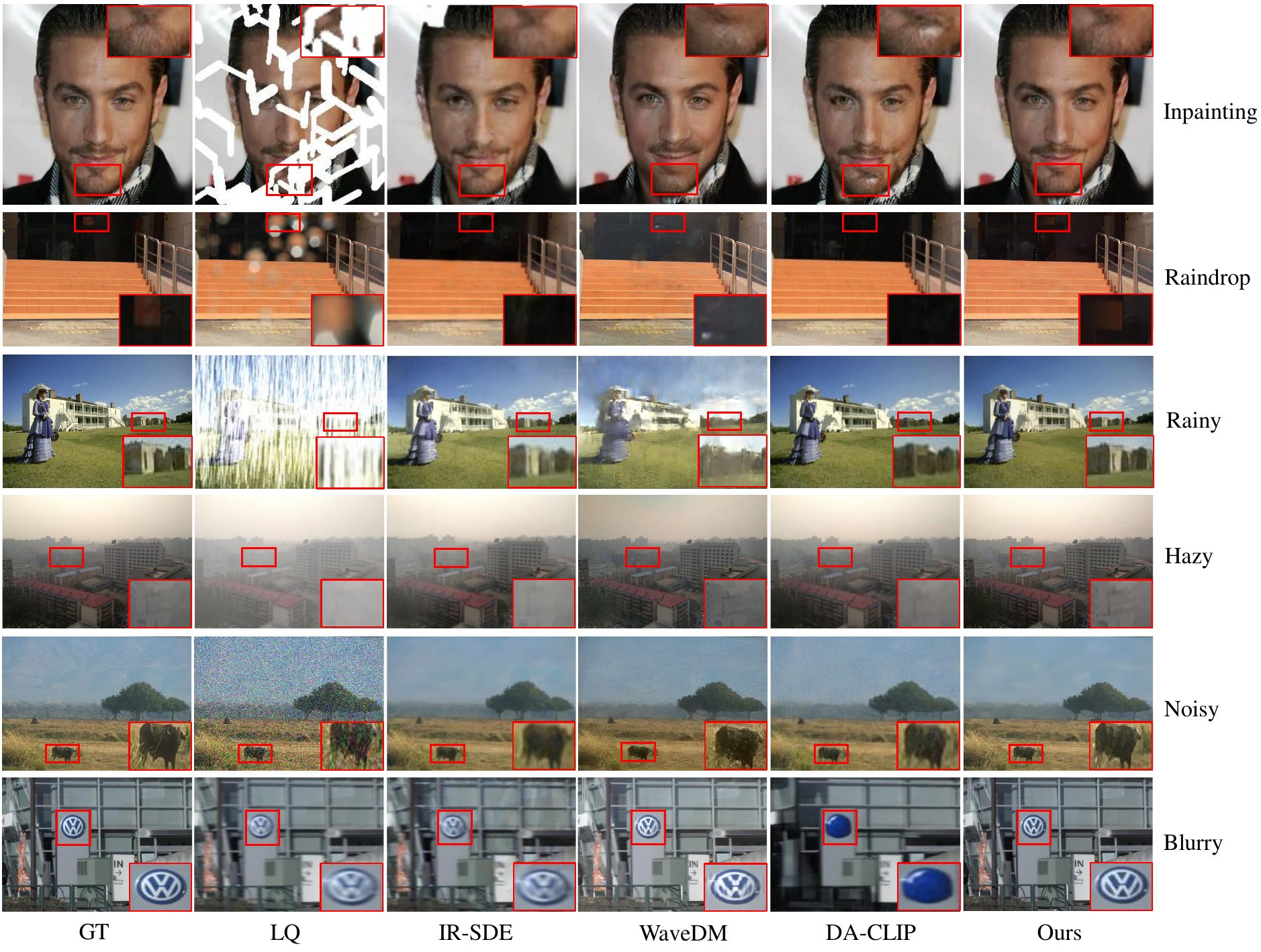}  
        \caption{Comparison of our method with other state-of-the-art methods in image restoration tasks with different degradation types. Best viewed by zooming in.}
        \label{fig:r2}
    \end{figure*}

{\bf Comparison Methods.} For all tasks, due to the specificity of the training, we focus more on the comparison with the current state-of-the-art methods for the specific task, such as 1) DBSN \cite{wu2020unpaired}, DCDicL \cite{zheng2021deep}, Restormer \cite{zamir2022restormer}, WCDM \cite{jiang2023low} for image denoising; 2) RecDerain \cite{ren2020single}, MPRNet \cite{zamir2021multi}, IDT \cite{xiao2022image}, WSDformer \cite{zhang2024prior} for image deraining; 3) MTRNN \cite{park2020multi}, DID-Anet \cite{ma2021defocus}, MSDI-Net \cite{li2022learning}, DeepRFT \cite{mao2023intriguing} for image deblurring; 4) FFANet \cite{qin2020ffa}, MSFNet \cite{zhu2021multi}, the UME-Net \cite{sun2024unsupervised} and D4\_plus \cite{yang2023robust} for image dehazing. We also compare with IR-SDE \cite{luo2023image} and DA-CLIP \cite{luo2024controlling}, which are two advanced image restoration network architectures for state-of-the-art performance on multiple tasks. In addition, for the image Inpainting and image raindrop removal tasks, we compare them with four image restoration methods, respectively.

{\bf Quantitative Results.} Table \ref{tab:1} summarises the quantitative results for four different image restoration tasks. Our method is vastly superior in all recovery tasks, where we achieve the top two evaluation results for all tasks in the SSIM and perceptual evaluations, and set several state-of-the-art performances. In addition, we also achieved first place in the deblurring task and second place in the deraining and dehazing tasks in the PSNR evaluation. Compared to the baseline methods IR-SDE and DA-CLIP, our method overall improves the metrics results for all datasets, demonstrating better generalization performance and stability. 

In addition, Table \ref{tab:r1} demonstrates the performance in the image inpainting and image raindrop removal tasks. Compared to the three multitasking image restoration methods, we obtained the best performance in the image inpainting task for SSIM, LPIPS and FID evaluations, and acquired the second place in the PSNR evaluation. In the image raindrop removal task on the other hand, we obtained the best performance evaluation in PSNR and LPIPS and the second place in SSIM and FID evaluation. This further demonstrates the generalisation ability of our method for linear image restoration tasks as well as better machine-aware results.

{\bf Qualitative Results.} As shown in Fig. \ref{fig:3}, we first show a visual comparison of the four tasks of de-fogging, de-noising, de-blurring and de-raining with a wide range of method-specific experiments. It can be seen that our methods are clearly competitive in terms of details and overall visual effects. For example, in the de-fogging task, MSFNet and DA-CLIP cannot effectively remove white fog. In the deblurring task, we were able to further recover a clear visual effect. This further validates the effectiveness of CycleRDM to achieve a restoration effect that is more satisfying to human visual perception.

\begin{table*}[t]
\renewcommand\arraystretch{1.05}
    \caption{Quantitative comparison of our method with other SOTA methods on low-light image enhancement image enhancement tasks. The best and second performance are marked in {\textcolor[HTML]{FF0000}{red}} and {\color[HTML]{00B0F0}{blue}}, respectively. } 
    \label{Tab:2}
      \centering
        \resizebox{!}{3.4cm}{
\begin{tabular}{l|c|cccccc|cccccc}
\hline
                          &                             & \multicolumn{6}{c|}{LOLv1}                                                                                                                                                                 & \multicolumn{6}{c}{LOLv2\_Real}                                                                                                                                                          \\ \cline{3-14} 
\multirow{-2}{*}{Methods} & \multirow{-2}{*}{Reference} & PSNR↑                          & SSIM↑                         & LPIPS↓                        & FID↓                           & MUSIQ↑                         & VIF↑                          & PSNR↑                          & SSIM↑                         & LPIPS↓                        & FID↓                           & MUSIQ↑                         & VIF↑                          \\ \hline
SNRnet \cite{xu2022snr}                   & CVPR'22                     & 24.309                        & 0.841                        & 0.262                        & 56.467                        & 65.225                        & 0.488                        & 21.480                        & 0.849                        & 0.237                        & 54.532                        & 62.506                        & 0.536                        \\
NeRco \cite{yang2023implicit}                    & ICCV'23                     & 22.946                        & 0.785                        & 0.311                        & 76.727                        & 68.964                        & 0.359                        & 25.172                        & 0.785                        & 0.338                        & 84.534                        & 66.159                        & 0.348                        \\
CLIP-LIT \cite{liang2023iterative}                 & ICCV'23                     & 12.394                        & 0.493                        & 0.397                        & 108.739                       & 57.099                        & 0.434                        & 15.262                        & 0.601                        & 0.398                        & 100.459                       & 55.904                        & 0.456                        \\
GDP  \cite{fei2023generative}                     & CVPR'23                     & 15.904                        & 0.540                        & 0.431                        & 112.363                       & 60.585                        & 0.380                        & 14.290                        & 0.493                        & 0.435                        & 102.416                       & 58.381                        & 0.424                        \\
UHDFour \cite{li2023embedding}                  & ICLR'23                     & 23.095                        & 0.822                        & 0.259                        & 56.912                        & 59.019                        & 0.487                        & 21.785                        & 0.854                        & 0.291                        & 60.849                        & 66.023                        & 0.536                        \\
GSAD \cite{hou2024global}                     & NeurIPS'23                  & {\color[HTML]{FF0000} 27.629} & {\color[HTML]{FF0000} 0.876} & {\color[HTML]{00B0F0} 0.188} & 43.659                        & {\color[HTML]{00B0F0} 72.479} & {\color[HTML]{00B0F0} 0.547} & 28.805                        & {\color[HTML]{00B0F0} 0.894} & 0.201                        & 41.456                        & {\color[HTML]{00B0F0} 69.139} & 0.532                        \\
WCDM \cite{jiang2023low}                     & TOG'23                      & {\color[HTML]{00B0F0} 26.316} & 0.844                        & 0.219                        & 48.037                        & 66.896                        & 0.462                        & {\color[HTML]{00B0F0} 28.875} & 0.874                        & 0.203                        & 45.395                        & 64.389                        & 0.508                        \\
FourierDiff \cite{lv2024fourier}                      & CVPR'24                    & 17.560                        & 0.607                        & 0.359                        & 77.768                        & 57.289                        & 0.485                        & 17.304                        & 0.783                        & 0.303                        & 63.499                        & 54.299                        & 0.503                        \\
LightenDiffusion \cite{jiang2024lightendiffusion}                  & ECCV'24                     & 20.188                        & 0.814                        & 0.316                        &85.930 & 57.952                        & 0.463                        & 22.443 & 0.867                        & 0.305 & 75.582 & 57.952                        & 0.508 \\
DA-CLIP \cite{luo2024controlling}                  & ICLR'24                     & 23.528                        & 0.810                        & 0.204                        & {\color[HTML]{FF0000} 34.852} & 70.587                        & 0.497                        & {\color[HTML]{FF0000} 31.009} & 0.851                        & {\color[HTML]{00B0F0} 0.187} & {\color[HTML]{FF0000} 27.962} & 65.254                        & 0.546 \\
UPT-Flow \cite{xu2025upt}                & PR’25                     & 20.644                       & 0.862                       & 0.215                        & 48.926                        & 60.768                        & 0.507                        & 25.056                        & 0.889                        & 0.231                        & 50.757                        & 57.952                       & {\color[HTML]{00B0F0} 0.556}                        \\ 
Ours                      & -                            & 24.423                        & {\color[HTML]{00B0F0} 0.865} & {\color[HTML]{FF0000} 0.179} & {\color[HTML]{00B0F0} 41.128} & {\color[HTML]{FF0000} 72.663} & {\color[HTML]{FF0000} 0.552} & 25.893                        & {\color[HTML]{FF0000} 0.901} & {\color[HTML]{FF0000} 0.158} & {\color[HTML]{00B0F0} 32.368} & {\color[HTML]{FF0000} 69.214} & {\color[HTML]{FF0000} 0.606} \\ \hline
\end{tabular}
        }
\end{table*}

\begin{table*}[ht!]
    \caption{Quantitative comparison of our method with other state-of-the-art methods between underwater image enhancement and backlight image enhancement tasks.} 
    \label{Tab:e2}
    \begin{subtable}{.65\linewidth}
      \centering
        \resizebox{!}{2.95cm}{
\begin{tabular}{l|c|cccccc}
\hline
                          &                             & \multicolumn{6}{c}{LSUI}                                                                                                                                                                   \\ \cline{3-8} 
\multirow{-2}{*}{Methods} & \multirow{-2}{*}{Reference} & PSNR↑                          & SSIM↑                         & LPIPS↓                        & FID↓                           & MUSIQ↑                         & VIF↑                          \\ \hline
TACL \cite{liu2022twin}                     & TIP'22                      & 21.150                        & 0.781                        & 0.238                        & 56.381                        & 44.176                        & 0.525                        \\
PUIE-Net \cite{fu2022uncertainty}                 & ECCV'22                     & 21.782                        & 0.813                        & 0.255                        & 51.698                        & 43.328                        & 0.579                        \\
USUIR \cite{fu2022unsupervised}                    & AAAI'22                     & 19.683                        & 0.783                        & 0.298                        & 59.291                        & 43.910                        & 0.564                        \\
Semi-UIR \cite{huang2023contrastive}                 & CVPR'23                     & 22.054                        & 0.814                        & 0.194                        & 50.115                        & {\color[HTML]{00B0F0} 51.311} & {\color[HTML]{00B0F0} 0.645} \\
PUGAN \cite{cong2023pugan}                    & TIP'23                      & {\color[HTML]{00B0F0} 22.507} & 0.816                        & 0.215                        & 51.622                        & 45.481                        & 0.556                        \\
GUPDM \cite{mu2023generalized}                    & ACM MM'23                   & 22.372                        & 0.834 & {\color[HTML]{00B0F0} 0.167} & {\color[HTML]{00B0F0} 42.313} & 50.731 & 0.617                        \\
WCDM  \cite{jiang2023low}                    & TOG'23                      & 22.286       & {\color[HTML]{00B0F0} 0.837}                             &0.197                              &  51.238                             & 48.573                              & 0.611                             \\
DA-CLIP \cite{luo2024controlling}                  & ICLR'24                     & 21.342                        & 0.799                        & 0.209                        & 49.577                        & 47.947                        & 0.578                        \\
Ours                & -     & {\color[HTML]{FF0000} 22.737} & {\color[HTML]{FF0000} 0.853} & {\color[HTML]{FF0000} 0.141} & {\color[HTML]{FF0000} 35.467} & {\color[HTML]{FF0000} 51.682} & {\color[HTML]{FF0000} 0.666} \\ \hline
\end{tabular}
        }
        \caption{Underwater image enhancement on the LSUI dataset}

    \end{subtable}%
    \begin{subtable}{.35\linewidth}
      \centering

        \resizebox{!}{2.95cm}{
\begin{tabular}{l|c|c}
\hline
                          &                             & Backlit300                    \\ \cline{3-3} 
\multirow{-2}{*}{Methods} & \multirow{-2}{*}{Reference} & MUSIQ↑                         \\ \hline
SNRnet \cite{xu2022snr}                   & CVPR'22                     & 53.734                        \\
CLIP-LIT \cite{liang2023iterative}                 & ICCV'23                     & {\color[HTML]{00B0F0} 63.141} \\
GDP  \cite{fei2023generative}                     & CVPR'23                     & 54.578                        \\
UHDFour \cite{li2023embedding}                  & ICLR'23                     & 59.587                        \\
WCDM  \cite{jiang2023low}                    & TOG'23                      & 59.420                        \\
IR-SDE  \cite{luo2023image}                  & ICML'23                      & 59.248                       \\
GSAD   \cite{hou2024global}                   & NeurIPS'23                  & 58.969                        \\
FourierDiff \cite{lv2024fourier}              & CVPR'24                    & 62.259                       \\
DA-CLIP  \cite{luo2024controlling}                 & ICLR'24                     & 60.178                        \\
Ours                      & -                            & {\color[HTML]{FF0000} 63.846} \\ \hline
\end{tabular}
        }
        \caption{Backlight Enhancement on the Backlit300 dataset.}

    \end{subtable} 
\end{table*}

Also, we show a visual comparison with three multitasking methods IR-SDE, WaveDM and DA-CLIP on all image restoration tasks. As shown in Fig. \ref{fig:r2}, CycleRDM achieves perceptually oriented recovery results compared to the baseline methods. For example, IR-SDE fails to remove and recover specific regions efficiently in the image defogging and image inpainting tasks.DA-CLIP distorts some details in the image inpainting and deblurring tasks, which leads to unsatisfactory visual perceptions. This indicates that CycleRDM achieves effective reconstruction of detail regions in multi-stage refinement.
\begin{table}[t]
\caption{Quantitative comparison of our method with other state-of-the-art methods in the unpaired datasets DICM and LIME. The best and second performance are marked in {\textcolor[HTML]{FF0000}{red}} and {\color[HTML]{00B0F0}{blue}}, respectively. }
\centering
\scalebox{0.83}{
\begin{tabular}{l|c|c|c}
\hline
                          &                             & DICM                          & LIME                          \\ \cline{3-4} 
\multirow{-2}{*}{Methods} & \multirow{-2}{*}{Reference} & MUSIQ↑                         & MUSIQ↑                         \\ \hline
SNRNet \cite{xu2022snr}                   & CVPR'22                     & 53.594                        & 52.165                        \\
CLIP-Lit \cite{liang2023iterative}                 & ICCV'23                     & 63.637                        & 62.865                        \\
NeRCo \cite{yang2023implicit}                    & ICCV'23                     & 63.878                        & {\color[HTML]{00B0F0} 66.122} \\
UHDFour  \cite{li2023embedding}                 & ICLR'23                     & 59.238                        & 58.125                        \\
GDP \cite{fei2023generative}                      & CVPR'23                     & 55.134                        & 57.320                        \\
GSAD \cite{hou2024global}                     & NeurIPS'23                  & {\color[HTML]{00B0F0} 65.371} & 59.945                        \\
IR-SDE \cite{luo2023image}   & ICML'23   & 61.279  & 55.134 \\
WCDM    \cite{jiang2023low}                  & TOG'23                      & 58.264                        & 57.617                        \\
FourierDiff \cite{lv2024fourier}                      & CVPR'24                     & 60.882                        & 61.012                        \\
DA-CLIP  \cite{luo2024controlling}                 & ICLR'24                     & 63.559                        & 62.516                      \\
UPT-Flow \cite{xu2025upt}                & PR’25                     & 62.251                      & 65.564                      \\
Ours                      & -                           & {\color[HTML]{FF0000} 65.581} & {\color[HTML]{FF0000} 66.253} \\ \hline
\end{tabular}
}
\label{tab:ee2}
\end{table}

\begin{figure}[ht!]
        \centering
        \includegraphics[height=0.7\linewidth,width=\linewidth]{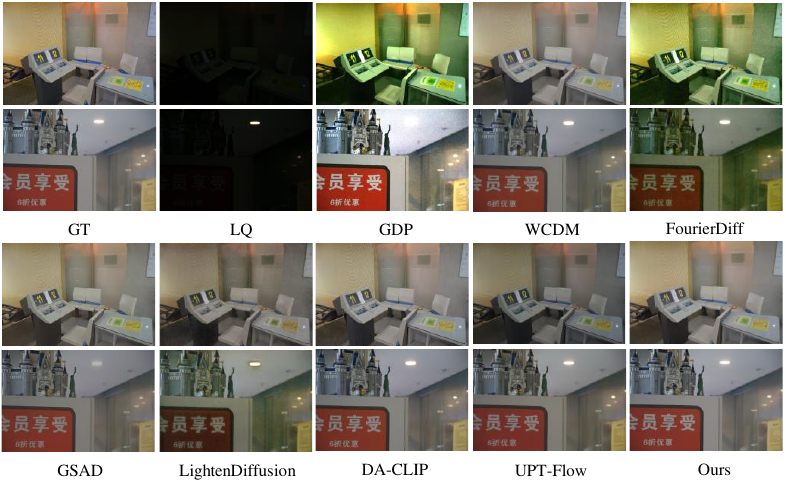}  
        \caption{Comparison of our method with competing methods on low light image enhancement task. Best viewed by zooming in.}
        \label{fig:4}
    \end{figure}

\begin{figure*}[ht!]
        \centering
        \includegraphics[height=0.43\textwidth,width=\textwidth]{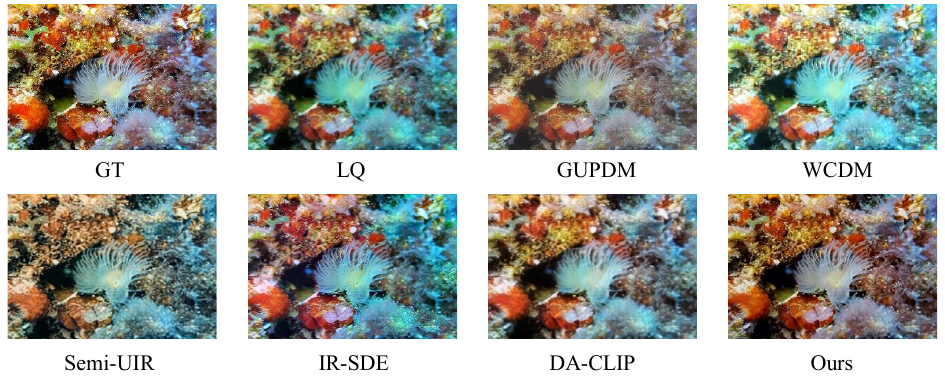}  
        \caption{Comparison of our method with other state-of-the-art methods in underwater image enhancement tasks.}
        \label{fig:f4}
    \end{figure*}   
 \begin{figure*}[h!]
        \centering
        \includegraphics[height=0.6\textwidth,width=\textwidth]{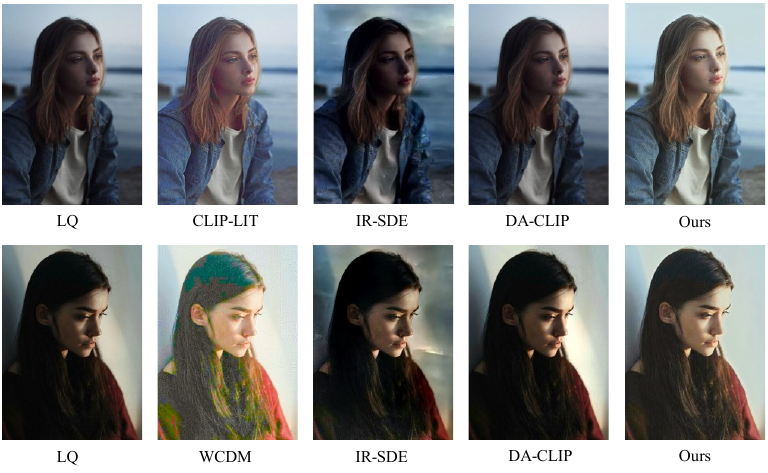}  
        \caption{Comparison of our method with other state-of-the-art methods in backlit image enhancement.}
        \label{fig:f3}
    \end{figure*}

\subsection{Image Enhancement Tasks}
Encouraged by the excellent performance on the image recovery task, we further evaluate the generalization ability of CycleRDM for more challenging non-linear image enhancement tasks. We demonstrate three enhancement tasks: low-light image enhancement on the LOLv1 \cite{wei2018deep} and LOLv2\_Real \cite{yang2021sparse} datasets, underwater image enhancement on the LSUI dataset \cite{peng2023u}, and backlight image enhancement on the BackLit300 unpaired dataset \cite{liang2023iterative}. In addition, we add MUSIQ \cite{ke2021musiq} and visual information fidelity(VIF) \cite{sheikh2005live} metrics to evaluate the model performance further. For the unpaired dataset BackLit300, we only use MUSIQ for evaluation.

{\bf Comparison Methods.}To further validate the model performance, we conducted a comprehensive comparison of three SOTA methods for specific image enhancement tasks. All methods are listed below: SNRNet \cite{xu2022snr}, NeRco \cite{yang2023implicit}, CLIP-LIT \cite{liang2023iterative}, GDP \cite{fei2023generative}, UHDFour \cite{li2023embedding}, GSAD \cite{hou2024global}, FourierDiff \cite{lv2024fourier}, LightenDiffusion \cite{jiang2024lightendiffusion}, TACL \cite{liu2022twin}, PUIE-Net \cite{fu2022uncertainty}, USUIR \cite{fu2022unsupervised}, Semi-UIR \cite{huang2023contrastive}, PUGAN\cite{cong2023pugan}, GUPDM \cite{mu2023generalized}, WCDM \cite{jiang2023low}, DA-CLIP \cite{luo2024controlling}, UPT-Flow \cite{xu2025upt}.

{\bf Quantitative Results.} As shown in Table \ref{Tab:2}, except for the PSNR evaluation, our method obtains the top two evaluation performances on all datasets for other metrics, and in particular for the perception metric MUSIQ, LPIPS and VIF, we redefine the state-of-the-art performance for all datasets. More importantly, the competing methods are focused on augmentation tasks and do not have superior task expansion capabilities. Table \ref{Tab:e2} shows the performance comparison in the underwater image enhancement and backlight image enhancement tasks. Despite the fact that CycleRDM uses only a small amount of training data, it still obtains excellent performance, obtaining the best evaluations for both tasks. In addition, as shown in Table \ref{tab:ee2}, to further validate the model generalisation ability, we tested it on the unpaired datasets LIME \cite{guo2016lime} and DICM \cite{lee2013contrast}. It can be clearly seen that CycleRDM acquires the optimal performance, which fully verifies the excellent processing capability of CycleRDM for different scenarios.

{\bf Qualitative Results.} Fig. \ref{fig:4}, Fig. \ref{fig:f4} and Fig. \ref{fig:f3} show visual comparisons in the low-light image enhancement, underwater image enhancement and backlit image enhancement tasks, respectively. Among them, in Fig. \ref{fig:4}, we find that FourierDiff and WCDM have some colour distortion.LightenDiffusion and UPT-Flow do not achieve good enhancement. enhancement. And CyclerRDM can reach an effective balance between colour and luminance as GSAD, so as to obtain a satisfactory visual perception. For the underwater For the underwater image enhancement task, in Fig. \ref{fig:f4} we can easily see that the competing methods do not effectively remove the redundant features and colours of the underwater environment. In contrast, CycleRDM can recover clean and clear images and can achieve excellent detail processing. In addition, from Fig. \ref{fig:f3} we see that IR-SDE and DA-CLIP are limited by their insensitivity to the backlighting task, which results in the model not being able to show effective generalisation ability in this task. By comparison, we further confirm the effectiveness of CycleRDM in image enhancement tasks, and we can perfect the unification of linear and blind tasks compared to other single models.

\begin{figure*}[t!]
        \centering
        \includegraphics[height=0.245\textwidth,width=\textwidth]{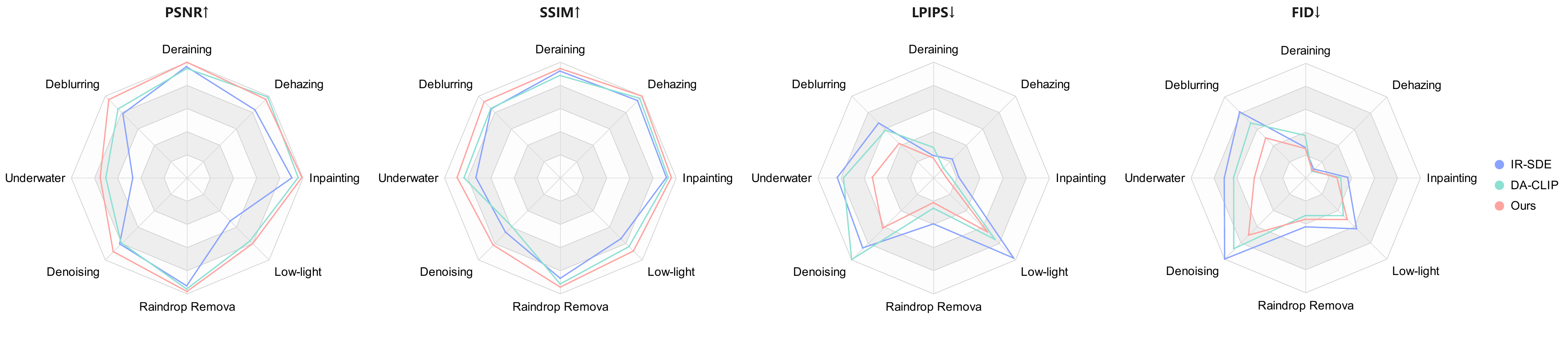}  
        \caption{Our method is compared with IR-SDE and DA-CLIP for uniform image recovery. Each radar plot reports results for eight different degradation types for a particular metric. For the perceptual metrics LPIPS and FID, lower values are better.}
        \label{fig:1}
    \end{figure*}

\begin{table*}[ht!]
    \centering
        \caption{Here, we also provide a more detailed comparison of our method with other unified image restoration method baselines. The best performance for each metric is \textbf{bolded}. }
    \begin{subtable}[t]{0.495\linewidth}
\resizebox{!}{3.3cm}{
\begin{tabular}{l|ccc}
\hline
Task       & IR-SDE & DA-CLIP         & Ours            \\ \hline
Dehazing   & 25.250 & \textbf{30.062} & 29.202          \\
Denoising  & 24.821 & 24.333          & \textbf{27.424} \\
Deblurring & 23.723 & 25.631          & \textbf{29.056} \\
Deraining  & 29.097 & 28.756          & \textbf{30.407} \\
Inpainting & 27.577 & 29.277          & \textbf{30.292} \\
Raindrop   & 28.491 & 29.305          & \textbf{29.857} \\
Low-light  & 16.072 & 23.528          & \textbf{24.423} \\
Underwater & 14.284 & 21.342          & \textbf{22.737} \\
Average    & 23.664 & 26.529          & \textbf{27.925} \\ \hline
\end{tabular}
}
        \caption{Comparison of our method with the baseline method in PSNR.}
    \end{subtable}
    \begin{subtable}[t]{0.495\linewidth}
\resizebox{!}{3.3cm}{
\begin{tabular}{l|ccc}
\hline
Task       & IR-SDE & DA-CLIP & Ours           \\ \hline
Dehazing   & 0.908  & 0.935   & \textbf{0.961} \\
Denoising  & 0.641  & 0.571   & \textbf{0.789} \\
Deblurring & 0.806  & 0.812   & \textbf{0.889} \\
Deraining  & 0.886  & 0.849   & \textbf{0.908} \\
Inpainting & 0.884  & 0.901   & \textbf{0.926} \\
Raindrop   & 0.836  & 0.882   & \textbf{0.908} \\
Low-light  & 0.719  & 0.811   & \textbf{0.865} \\
Underwater & 0.699  & 0.799   & \textbf{0.853} \\
Average    & 0.797  & 0.820   & \textbf{0.887} \\ \hline
\end{tabular}
}
        \caption{Comparison of our method with the baseline method in SSIM.}
    \end{subtable}
        \begin{subtable}[t]{0.495\linewidth}
        \resizebox{!}{3.3cm}{
\begin{tabular}{l|ccc}
\hline
Task       & IR-SDE         & DA-CLIP & Ours           \\ \hline
Dehazing   & 0.062          & 0.033   & \textbf{0.026} \\
Denoising  & 0.232          & 0.269   & \textbf{0.166} \\
Deblurring & 0.179          & 0.156   & \textbf{0.113} \\
Deraining  & \textbf{0.050} & 0.071   & \textbf{0.050} \\
Inpainting & 0.061          & 0.042   & \textbf{0.039} \\
Raindrop   & 0.107          & 0.061   & \textbf{0.058} \\
Low-light  & 0.231          & 0.195   & \textbf{0.168} \\
Underwater & 0.268          & 0.209   & \textbf{0.141} \\
Average    & 0.149          & 0.130   & \textbf{0.095} \\ \hline
\end{tabular}
}
        \caption{Comparison of our method with the baseline method in LPIPS.}
    \end{subtable}
        \begin{subtable}[t]{0.495\linewidth}
\resizebox{!}{3.3cm}{
\begin{tabular}{l|ccc}
\hline
Task       & IR-SDE & DA-CLIP         & Ours            \\ \hline
Dehazing   & 8.330  & \textbf{5.341}  & 7.095           \\
Denoising  & 79.380 & 69.908          & \textbf{55.947} \\
Deblurring & 64.506 & 53.660          & \textbf{38.643} \\
Deraining  & 20.869 & 29.214          & \textbf{20.318} \\
Inpainting & 29.605 & 22.684          & \textbf{22.053} \\
Raindrop   & 34.221 & \textbf{22.379} & 28.879          \\
Low-light  & 52.466 & \textbf{34.852} & 41.128          \\
Underwater & 54.131 & 49.577          & \textbf{35.467} \\
Average    & 42.939 & 35.952          & \textbf{31.191} \\ \hline
\end{tabular}
}
        \caption{Comparison of our method with the baseline method in FID.}
    \end{subtable}
    \label{tab:8}
\end{table*}


\begin{table}[ht!]
\caption{Comparison of the average results over all different datasets on the unified image restoration task. }
\centering
\scalebox{0.8}{
\begin{tabular}{l|c|cccc}
\hline
Methods & Reference & PSNR↑            & SSIM↑           & LPIPS↓          & FID↓             \\ \hline
IR-SDE  & ICML'23   & 23.664         & 0.797          & 0.149          & 42.939          \\
DA-CLIP & ICLR'24   & 26.529          & 0.820          & 0.130          & 35.952          \\
Ours    & -         & \textbf{27.925} &\textbf{0.887}& \textbf{0.095} & \textbf{31.191} \\ \hline
\end{tabular}
}
\label{tab:38}
\end{table}

\subsection{Detailed Performance Analysis}
This subsection compares our method with other unified image restoration methods in detail. As shown in Fig. \ref{fig:1} and Table \ref{tab:8}, we summarise in detail the metrics comparison with the two baseline methods, IR-SDE and DA-CLIP, on 8 tasks. In addition, as shown in Table \ref{tab:38}, we computed the average metric results of our method and the baseline method on all tasks separately to show more overall performance. It is intuitively clear that DA-CLIP, although it can be optimised to some extent for various degradation tasks, this optimisation does not compete strongly with the current state-of-the-art task-specific methods and, in particular, has significant drawbacks in terms of distortion metrics. Meanwhile, compared with IR-SDE based on the diffusion model, attributed to the multi-level refinement of image restoration, we have greatly narrowed the performance gap between the various tasks, and achieved a perfect balance of the diffusion model between the image restoration and enhancement tasks, which leads to a more stable restoration result of the model. Among them, compared with IR-SDE and DA-CLIP, we have risen \textbf{4.261 dB and 1.396 dB} in PSNR evaluation, respectively. in SSIM, we have \textbf{0.09 and 0.067} rise. For the distortion metrics LPIPS and FID, we also have a significant lead with \textbf{0.054/0.035 and 11.748/4.761} uplifts, respectively.

\subsection{Ablation Study}
In this subsection, we conduct a series of ablation studies to measure the impact of the different component configurations employed in our approach. Specific details are given below:

{\bf The Effectiveness of the Multi-stage inference process.} ablation experiments on a low-light enhancement task, we verified their effectiveness. As shown in Table \ref{tab:33}, where \#1 denotes single-stage inference using only Stage1, \#2 does not perform the final wavelet calibration process, \#3 does not perform the rough normal-domain-to-normal-domain learning process, and \#4 uses the full setup. We can see that both \#2 and \#3 obtain significant performance gains compared to single-stage inference \#1, especially for the PSNR evaluation, which obtains 2.293 dB and 3.031 dB, respectively, but degrades for the perceptual metric FID. To further refine the degraded domain, we applied the default setting of configuration \#4. Among the four model configurations, \#4 achieved the best evaluation results, demonstrating a notable improvement. Specifically, it enhanced the PSNR metric by an additional 1.246 dB compared to configuration \#3, while simultaneously alleviating the perceptual metric degradation in FID. These results validate that the refined multi-stage degradation priors, in conjunction with the calibration process, significantly enhanced the mapping quality from the degraded domain to the normal domain. This underscores the effectiveness of the proposed CycleRDM framework.

\begin{table}[t]
\centering
\caption{Ablation studies of multi-stage inference process.}
\scalebox{0.98}{
\begin{tabular}{c|ccc|cccc}
\hline
index & Stage 1 & Stage 2 & Stage 3 & PSNR↑   & SSIM↑  & LPIPS↓  & FID↓    \\ \hline
\#1   & \usym{1F5F8}      & \usym{2613}       & \usym{2613}       & 20.146 & 0.835 & 0.201 & 41.059 \\
\#2   & \usym{1F5F8}      & \usym{1F5F8}      & \usym{2613}       & 22.439 & 0.848 & 0.198 & 46.213 \\
\#3   & \usym{1F5F8}      & \usym{2613}       & \usym{1F5F8}      & 23.177 & 0.854 & 0.191 & 46.998 \\
\#4   & \usym{1F5F8}      & \usym{1F5F8}      & \usym{1F5F8}      & \textbf{24.423} & \textbf{0.865} & \textbf{0.179} & \textbf{41.128} \\ \hline
\end{tabular}
}
\label{tab:33}
\end{table}

\begin{table}[t]
\centering
\caption{Ablation studies of the Feature Gain Module.}
\scalebox{0.98}{
\begin{tabular}{c|cccc}
\hline
version       & PSNR↑   & SSIM↑  & LPIPS↓  & FID↓            \\ \hline
CycleRDM & 27.655          & 0.952          & 0.027          & \textbf{6.906}          \\
CycleRDM+FGM & \textbf{29.202} & \textbf{0.960} & \textbf{0.026} & 7.095 \\ \hline
\end{tabular}
}
\label{tab:44}
\end{table}

\begin{table}[t!]
\centering
\caption{Ablation studies of the loss function terms.}
\scalebox{0.98}{
\begin{tabular}{l|cccc}
\hline
 Setting            & PSNR↑            & SSIM↑           & LPIPS↓          & FID↓                                   \\ \hline
without $\mathcal{L}_{diff}$ & 28.131                                 & 0.895                                & 0.098                                 & 82.892                                 \\
without  $\mathcal{L}_{content}$    & 25.478                                       &0.853                                       &0.101                                       &68.103                                        \\
without $\mathcal{L}_{fre} $ & 28.979                                 & 0.902                                 & 0.065                                 & 35.151                                 \\
without   $\mathcal{L}_{clip}$   & 28.305                                       & 0.900                                      &0.061                                       & 31.205                                       \\ \hline
Default      & { \textbf{29.857}} & {\textbf{0.908}} & {\textbf{0.058}} & {\textbf{28.879}} \\ \hline
\end{tabular}
}
\label{tab:88}
\end{table}

{\bf The Effectiveness of the Feature Gain Module.} We use the image dehazing task for validation, as shown in Table \ref{tab:44}, after adding the FGM on top of CycleRDM, the model improves in the distortion metrics, which indicates that the feature gain module can effectively remove the redundant features in the wavelet high-frequency information, which prompts the output results to be closer to the normal image. 

{\bf The Effectiveness of the Loss Function.} We also verify the validity of the proposed loss function in this subsection, where we perform the experiments by removing each component individually from the default settings, and the quantitative results are reported in Table \ref{tab:88}. As shown in row 1, deleting the diffusion loss $\mathcal{L}_{diff}$ leads to a significant reduction in the perceptual metrics, and the generative power of the diffusion model relies heavily on this component. The content loss $\mathcal{L}_{content}$ produces a noticeable change in the image generation results, especially in the distortion metrics, which can be improved by 4.379dB and 0.055 for PSNR and SSIM, respectively. The frequency domain perceptual loss $\mathcal{L}_{fre}$ and the multimodal text loss $\mathcal{L}_{clip}$ are intended to further aid in the reconstruction of image details and stabilize diffusion content generation, so its removal results in performance degradation. However, their degradation is significantly smaller than the diffusion loss and content loss. Thus, this reveals the importance of diffusion loss and content loss. From another perspective, the fact that content and diffusion loss should be combined with our proposed multi-stage diffusion inference training strategy illustrates their effectiveness. 

\section{LIMITATION AND FUTURE WORK}
Since degradation datasets mostly contain only a single degradation label for each image, our current model has not been trained to recover multiple degradations in the same scene. Although we have demonstrated the performance stability and generalization capabilities of CycleRDM when extended for degradation tasks, this limitation has prevented us from effectively exploring recovery capabilities for realistic mixed degradation scenes.

To address the above issues, in our future work, we will further explore the following aspects: 1) creating hybrid datasets containing multiple degraded scenes; 2) further exploring the real-time performance of the diffusion model for the task of unified image recovery; and 3) exploring further effective combinations of multimodal and diffusion models when dealing with degradation tasks.

\section{CONCLUSIONS}
In this paper, we propose CycleRDM, a novel framework designed to unify image restoration and enhancement tasks efficiently. CycleRDM employs a multilevel inference process to achieve high-quality mappings from degraded domains while progressively optimizing image details to ensure robust generalization. To further enhance its performance, we introduce a feature gain module that refines image restoration by effectively eliminating redundant features. Moreover, multimodal textual prompts are incorporated to positively guide the generation process, significantly boosting its recovery capabilities. Tested across nine tasks, CycleRDM delivers efficient and high-quality results using only a limited amount of training data, demonstrating its versatility and effectiveness in unifying image restoration and enhancement tasks.
\bibliography{PR}
\bibliographystyle{plain}


\end{document}